\documentclass[letterpaper]{article}

\usepackage{amssymb}
\usepackage{mathtools}
\usepackage{mathrsfs}
\newcommand{\defequal}{\mathrel{\overset{\makebox[0pt]{\mbox{\tiny def}}}{=}}}
\newcommand{\E}{\mathbb{E}} % Expectation
\usepackage{natbib}
\usepackage{alifeconf}
\usepackage{url}
\usepackage{booktabs}
\usepackage{hyperref}
\usepackage{cleveref}
\usepackage[bf]{caption}
\usepackage{subcaption}
\usepackage[space]{grffile}
\usepackage{xcolor}
\definecolor{dark-blue}{rgb}{0.05,0.1,0.75}
\hypersetup{
  colorlinks,
  linkcolor={dark-blue},
  citecolor={dark-blue},
  urlcolor={dark-blue}
}
\usepackage{todonotes}
\setuptodonotes{
  inline,
  linecolor=blue,
  backgroundcolor=blue,
  textcolor=white,
  bordercolor=blue,
}
\newcounter{num}

\newcommand{\M}{\mathcal{M}} % MDP
\newcommand{\A}{\mathcal{A}} % Actions
\newcommand{\X}{\mathcal{S}} % States
\usepackage{siunitx}
\usepackage{longtable} % tables that can span several pages
\graphicspath{{./resources}} % Specifies the directory where pictures are stored
\usepackage[indLines=false]{algpseudocodex}
\usepackage{algorithm}
\makeatletter
\algnewcommand\algorithmiconce{\textbf{Once}}%
\algdef{SE}[ONCE]{Once}{EndOnce}[1]{\algpx@startIndent\algpx@startCodeCommand\algorithmiconce\ #1\ \algorithmicdo%
}{\algpx@endIndent\algpx@startEndBlockCommand\algorithmicend\ \algorithmiconce}%
\pretocmd{\Once}{\algpx@endCodeCommand}{}{}
\algtext*{EndOnce}%
\apptocmd{\EndOnce}{\algpx@endIndent}{}{}%
\pretocmd{\EndOnce}{\algpx@endCodeCommand[1]}{}{}%

\algnewcommand\algorithmicWith{\textbf{With}}%
\algdef{SE}[WITH]{With}{EndWith}[1]{\algpx@startIndent\algpx@startCodeCommand\algorithmicWith\ #1\ \algorithmicdo%
}{\algpx@endIndent\algpx@startEndBlockCommand\algorithmicend\ \algorithmicWith}%
\pretocmd{\With}{\algpx@endCodeCommand}{}{}
\algtext*{EndWith}%
\apptocmd{\EndWith}{\algpx@endIndent}{}{}%
\pretocmd{\EndWith}{\algpx@endCodeCommand[1]}{}{}%
\makeatother
\title{Evolution of Rewards for Food and Motor Action by Simulating Birth and Death}
\author{
  Yuji Kanagawa$^{1}$ \and
  Kenji Doya$^{1}$ \\
  \mbox{}\\
  $^1$Okinawa Institute of Science and Technology Graduate University, Japan \\
  yuji.kanagawa@oist.jp
} % email of corresponding author

\begin{document}

\maketitle

\begin{abstract}
  The reward system is one of the fundamental drivers of animal behaviors and is critical for survival and reproduction. Despite its importance, the problem of how the reward system has evolved is underexplored. In this paper, we try to replicate the evolution of biologically plausible reward functions and investigate how environmental conditions affect evolved rewards' shape. For this purpose, we developed a population-based decentralized evolutionary simulation framework, where agents maintain their energy level to live longer and produce more children. Each agent inherits its reward function from its parent subject to mutation and learns to get rewards via reinforcement learning throughout its lifetime. Our results show that biologically reasonable positive rewards for food acquisition and negative rewards for motor action can evolve from randomly initialized ones. However, we also find that the rewards for motor action diverge into two modes: largely positive and slightly negative. The emergence of positive motor action rewards is surprising because it can make agents too active and inefficient in foraging. In environments with poor and poisonous foods, the evolution of rewards for less important foods tends to be unstable, while rewards for normal foods are still stable. These results demonstrate the usefulness of our simulation environment and energy-dependent birth and death model for further studies of the origin of reward systems.
\end{abstract}

\section{Introduction}\label{sec:intro}
% Problem, importantness, and specific question
Reward is a fundamental brain function that shapes our behavior through reward-based learning, or reinforcement learning (RL). Positive rewards encourage us to eat food and find mating partners, while negative rewards, such as pain and fatigue, help us protect ourselves. However, regardless of its importance to animal lives, the evolutionary process of such a reward system is underexplored. Reward signals should have evolved to help animals survive and reproduce offspring (e.g., by~\cite{schultzNeuronalRewardDecision2015}), but what kind of environmental conditions have influenced the evolution of varieties of rewards remains unclear.

\begin{figure}[t]
  \centering{}
  \includegraphics[width=7cm]{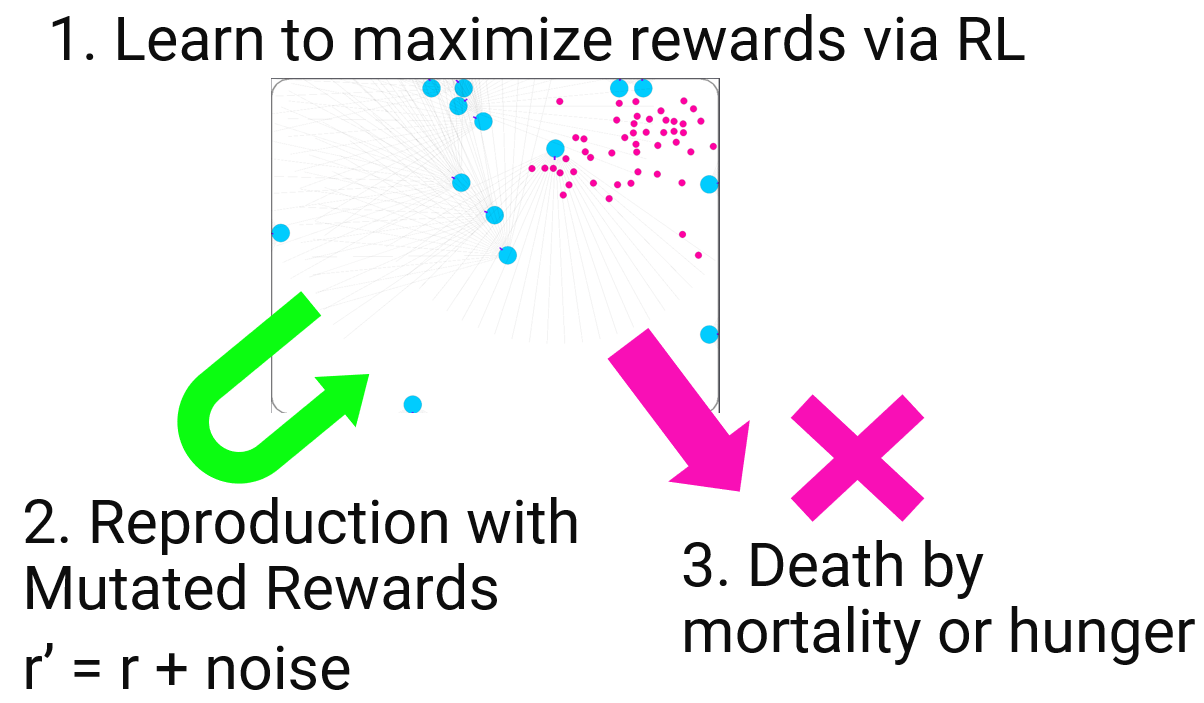}
  \caption{
    A schematic figure on our evolutionary framework.
    Agents learn to maximize the reward functions inherited from their parents.
    They reproduce children with mutated reward functions and die from aging or hunger.
  }\label{figure:scheme}
\end{figure}

% What we do
While biological studies in evolution are important, the difficulty of observing reward evolution in real animals motivates us to use artificial evolutionary simulation. Hence, we conduct an evolutionary simulation of RL agents to examine how different environmental conditions lead to different rewards. Each agent inherits a reward function from its parent and learns to get more rewards by RL during its lifetime. Contrary to previous studies on the evolution of learning \citep{hintonHowLearningCan1987,singhWhereRewardsCome2009} with centralized selection strategies, we design a distributed evolutionary model inspired by embodied evolution (EE) framework \citep{watsonEmbodiedEvolutionDistributing2002,bredecheEmbodiedEvolutionCollective2018}. In our model, birth and death for agents are evaluated independently based on their age and internal energy level, allowing the population change. With this approach, we expect rewards contributing to population growth to be selected. We show a schematic figure on our evolutionary framework in \Cref{figure:scheme}.

% Foraging environment
% Simulation
We implement our distributed evolution framework in a simple virtual environment where agents need to hunt foods to live longer and produce children based on 2D physics simulation. In this environment, we let agents evolve simple reward functions that map food intake and the magnitude of motor action to a scalar, which we expect to correspond to food pleasure and fatigue. Our results show that biologically reasonable reward functions with positive food rewards evolve from randomly initialized ones.  However, at the same time, we find that the rewards for motor action diverge into two modes: largely positive and slightly negative. We then confirm the influence of food density and distribution of the evolved rewards. We find that too much food supply stops evolution and that less food density and periodic relocation of foods lead to smaller rewards for motor action. Furthermore, we examine the evolution of rewards for poor and poisonous rewards, showing that rewards for less important foods are more unstable than those for normal foods. Overall, our results demonstrate the usefulness of our simulation environment and distributed evolution model for further studies. We then conclude the paper by discussing possible improvements and future works.

\section{Preliminaries and Related Works}\label{sec:related}
We follow the standard computational RL framework \citep{suttonReinforcementLearningIntroduction2018} based on the Markov decision process (MDP). MDP $\M{}$ consists of a tuple $(\X{}, \A{}, p, r, \gamma)$, where $\X{}$ is reward function, and $\gamma \in [0, 1]$ is the discount factor. A standard objective in MDP is the discounted cumulative return $G \defequal{} \sum_{t=0}^{\infty}\gamma^t R_{t}$, where $R_t$ is the reward received at time $t$. An RL agent has policy $\pi: \X \times \A \rightarrow [0, 1]$ and seeks to find the the optimal policy $\pi^{*}$ that maximizes $\E \left[G|\pi\right]$. The state-value function $V^\pi(s) \defequal \sum_{a \in \A} \pi(a|s) \left( r(s, a) + \gamma \sum_{s' \in \X} P(s'|s, a) V^\pi(s') \right)$ is often used in RL algorithms. Observation $o \in \Omega~(\Omega \subseteq \X)$ refers to a part of the state that an agent can observe.

Our reward model is inspired by the neuroscience of reward system \citep{schultzNeuronalRewardDecision2015, berridgePleasureSystemsBrain2015}. Notably, \citet{berridgeDissectingComponentsReward2009} argue that the brain reward system consists of three independent components: liking, wanting, and learning. In analogy with computational RL, liking corresponds to reward function, wanting corresponds to learned policy, and learning corresponds to learning state value $V$.

While there were previous attempts to evolve rewards in a single-agent setting \citep{singhWhereRewardsCome2009,niekumEvolutionRewardFunctions2011,zhengWhatCanLearned2020},
%While evolutionary robotics studies \citep{nolfiEvolutionaryRoboticsBiology2004} often employ a centralized selection scheme similar to genetic algorithm \citep{mitchellIntroductionGeneticAlgorithms1998},
%where highly evaluated elites are selected as parents. On the contrary,
we employ a distributed embodied evolution (EE) framework \citep{watsonEmbodiedEvolutionDistributing2002,bredecheEmbodiedEvolutionCollective2018} with many agents and population dynamics.
While the mainstream of evolutionary robotics studies \citep{nolfiEvolutionaryRoboticsBiology2004} employs a centralized elite selection, agents evolve locally following birth and death rules in EE. This method has the advantage that the evaluation of genetic traits depends on population dynamics and is more natural.

Our work is inspired by the series of studies \citep{elfwingBiologicallyInspiredEmbodied2005,elfwingDarwinianEmbodiedEvolution2011,elfwingEmergencePolymorphicMating2014}, which tried to evolve parameters related to RL through EE framework. Notably, \citet{elfwingDarwinianEmbodiedEvolution2011} evolved supplementary sharing rewards and parameters of RL agents.
Inspired by these works, we attempt to evolve the entire reward functions in our experiments.

\section{Simulation Model and Environment}\label{sec:method}

\begin{figure}[t]
  \centering{}
  \includegraphics[width=6cm]{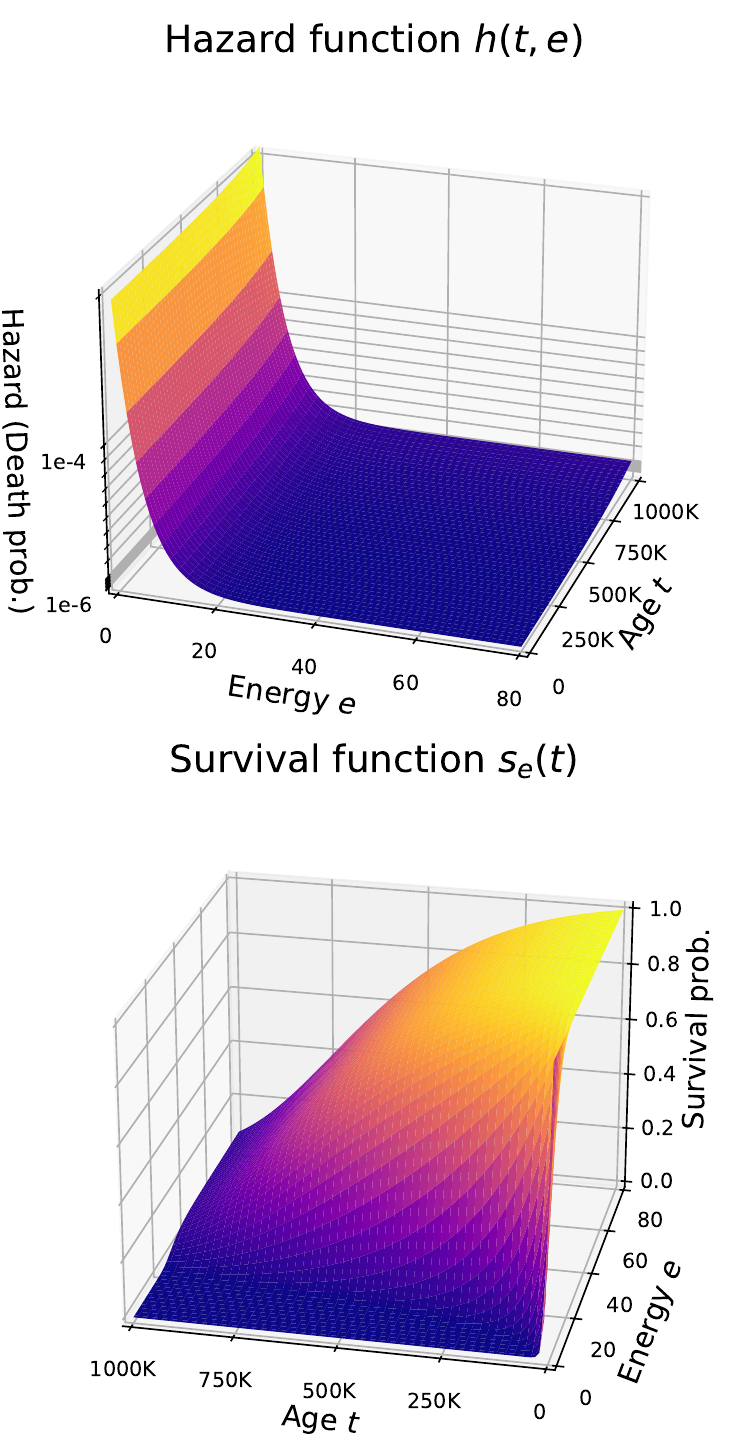}
  \caption{
    \textbf{Upper:} Hazard function $h(t)$ used in our experiments.
    \textbf{Lower:} Survival function $S(t)$ corresponding to the hazard function above.
  }\label{figure:hs}
\end{figure}

\paragraph{Based death and birth model with simple metabolism}
We employ a simple metabolic model similar to \citet{hamonEcoevolutionaryDynamicsNonepisodic2023}. Each agent maintains their energy level $e$, which increases by eating food and decreases by the basal metabolism and taking a motor action.
The death probability of the agent with energy level $e$ and age $t$ is modeled by the hazard function:
\begin{align}
  h(t, e) = \frac{\kappa_{h}}{1 + \alpha_{e}\exp(\beta_{he}e)} + \alpha_{t} \exp(\beta_{ht} t).
  \label{eq:h}
\end{align}
The first term in \Cref{eq:h} increases as energy levels decrease following a sigmoidal curve where $\kappa_{h}$ is the scale, $\beta_{he}$ is the slope, and $\alpha_{e}$ defines the shape when $e=0$. The latter term increases exponentially as the agent ages with the cale $\alpha_{t}$ and the slope $\beta_{ht}$, called Gompertz hazard model \citep{gompertzXXIVNatureFunction1825,kirkwoodDecipheringDeathCommentary2015}.
We show the shape of $h$ with parameters used in our experiments in the upper panel in \Cref{figure:hs}. The lower panel in \Cref{figure:hs} shows the survival function $s_{e}(t) = \exp (-\int_{0}^{t}(h(t, e)) dt)$ corresponding to $h$, which is the probability for an agent to survive to the age $t$ under the assumption that it keeps the same energy level $e$. We can see that the survival probability more sharply decays with aging when the energy level is low.

\begin{figure}[t]
  \centering{}
  \includegraphics[width=6cm]{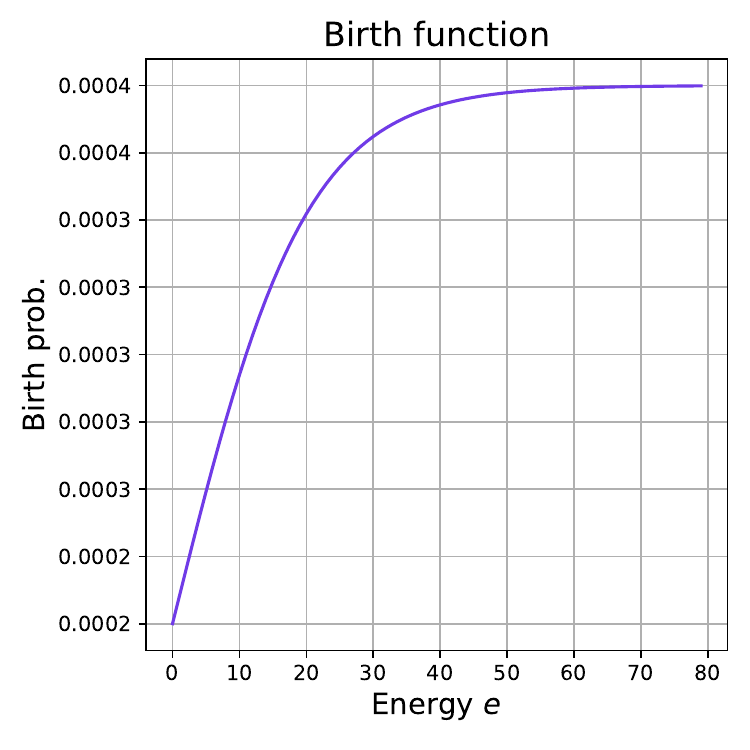}
  \caption{Birth function $b(e)$ used in our experiments.}\label{figure:birth}
\end{figure}

\begin{table}[t]
  \begin{subtable}[h]{0.45\columnwidth}
    \centering
    \begin{tabular}{cc}
      \toprule
      Parameter & Value \\
      \midrule
      $\kappa_{h}$ & 0.01 \\
      $\alpha_{e}$ & 0.02 \\
      $\beta_{he}$ & 0.2 \\
      $\alpha_{ht}$ & \num{2e-7} \\
      $\beta_{t}$ & \num{4e-6} \\
      \bottomrule
    \end{tabular}
  \end{subtable}
  \begin{subtable}[h]{0.45\columnwidth}
    \centering
    \begin{tabular}{cc}
      \toprule
      Parameter & Value \\
      \midrule
      $\kappa_{b}$ & \num{4e-4} \\
      $\beta_{b}$ & 0.1 \\
      \bottomrule
    \end{tabular}
  \end{subtable}
  \caption{Parameters of $h$ (left) and $b$ (right) used in our experiments.}\label{table:hb}
\end{table}

For simplicity, we employ an asexual reproduction model with the birth function, the probability for an agent with energy level $e$ to produce a child in a time step:
\begin{align}
 b(e) &= \frac{\kappa_{b}}{1 + \exp(\beta_{b}e)}.
 \label{eq:b}
\end{align}
The birth rate $b(e)$ increases with $e$ following a sigmoidal curve where $\kappa_{b}$ is the scale and $\beta_{b}$ is the slope.
\Cref{figure:birth} shows the shape of $b$ for the parameters used in our experiments. We show all paramters of $h$ and $b$ in \Cref{table:hb}. With this choice of parameters, we expect the maximum lifetime of an agent to be around \num{1e6} steps.

\paragraph{Environment}

\begin{figure}[t]
  \centering
  \includegraphics[width=6cm]{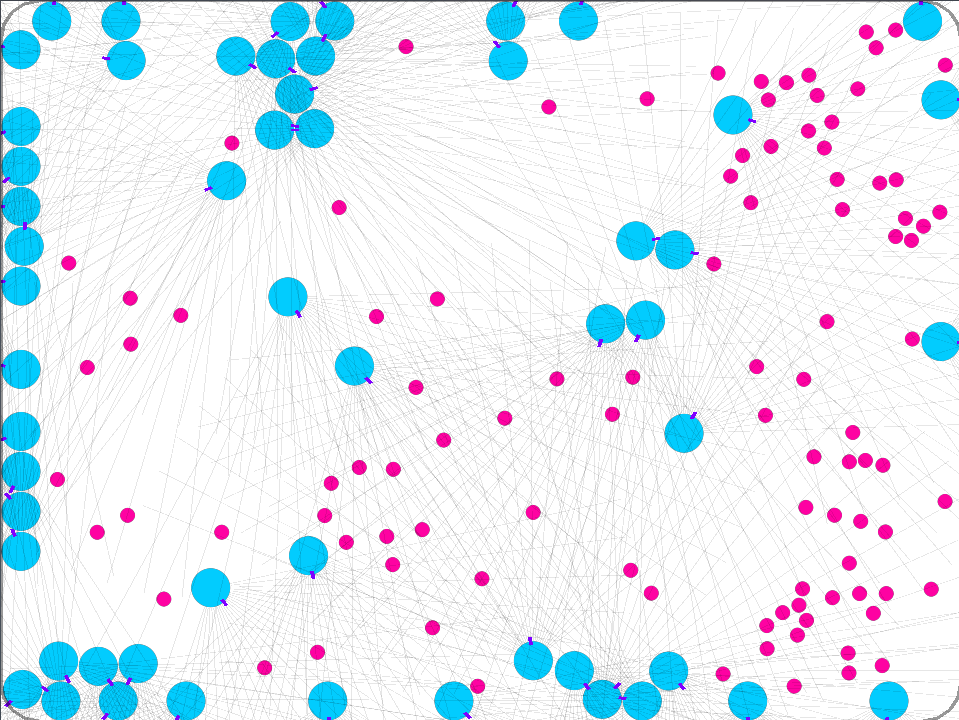}
  \caption{
    Simulation environment used in our experiments.
    Blue circles are agents, red circles are foods, and outer gray lines are walls.
    Thin gray lines around agents indicate distance sensors.
  }\label{figure:env}
\end{figure}

\begin{figure}[t]
  \begin{subfigure}[t]{4.5cm}
    \centering
    \includegraphics[width=4.5cm]{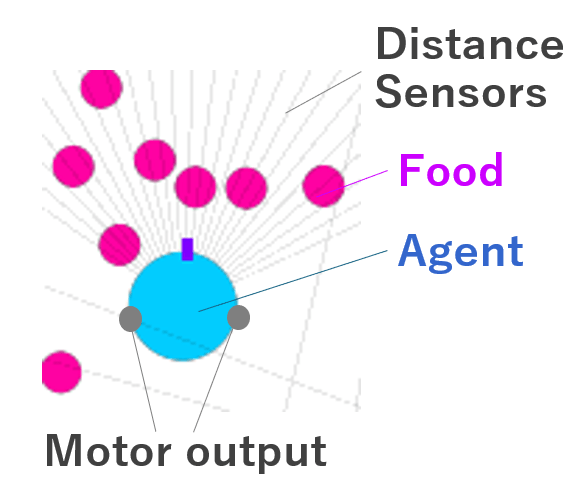}
  \end{subfigure}
  \begin{subfigure}[t]{3.5cm}
    \centering
    \includegraphics[width=3.5cm]{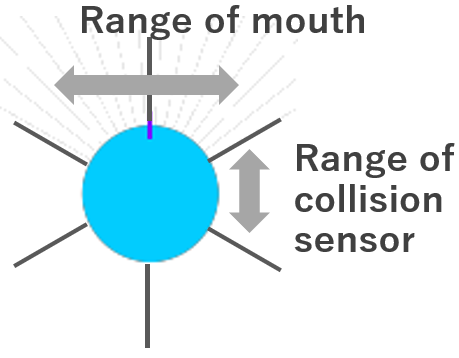}
  \end{subfigure}
  \caption{
    Description of the environment.
    The left figure shows an agent, foods, distance sensors, and the positions of motor outputs.
    The right figure shows the ranges of an agent's mouth and each collision sensor.
  }\label{figure:env-discr}
\end{figure}

We designed a continuous 2D environment shown in \Cref{figure:env}. Blue circles indicate agents, red circles indicate foods, and the outer gray lines are impassable walls. The environment is implemented by a 2D rigid-body physics simulation. Agents can move by producing driving forces on the left and right sides of the body, as shown in \Cref{figure:env-discr}. An agent has multiple range sensors in its front that can sense the type and distance to the closest object within $120$ degree. Sensible objects include foods, other agents, and walls. In environments with poor or poisonous foods, agents can sense them separately from normal foods. They can also sense the collision with each object and the approximate location of the collision with a resolution of $60$ degree, as shown in \Cref{figure:env-discr}.

% This eat-driven emergence is not consistent with the growth model below. Better report the actual food-producing process rather than theoretical approximation.
The agent can eat food by touching it within $120$ degrees of the front, as shown in the right figure in \Cref{figure:env-discr}. Eating one food gains energy $e_{\mathrm{food}}$. We use $e_{\mathrm{food}} = 1$ in our experiments. In experiments with nutrient-poor or poisonous foods, they have different energy gain $e_{\mathrm{poor}} = 0.2$ and $e_{\mathrm{poison}} = -0.4$. After foods are eaten, they are regenerated in a random place. To regulate the rate of food regeneration, we maintain the internal food number $n_{t}$ at time $t$ which follows a linear growth function: $n_{t + 1} = \textrm{min}(n_{t} + g - n_{t}^{\mathrm{eaten}}, n_{\mathrm{max}})$, where $g$ is the growth rate, $n_{\mathrm{max}}$ is the capacity of food, and $n_{t}^{\mathrm{eaten}}$ is the number of eaten foods at time $t$. Food is regenerated when the integer part of $n^{t}$ exceeds the number of foods in the environment. In most experiments, we use $n_{\mathrm{max}} = 100$ and $g = 0.02$.

The agent consumes energy via active and basic metabolism. At each step $t$, An agent $i$ losts energy $e_{\mathrm{act}} |a_{t}^{i}|$, where $|a_{t}^{i}|$ is the Euclid norm of the agent's motor output $a_{t}^{i}$ and $e_{\mathrm{act}}$ is a scaling coefficient. In addition, it also consumes energy $e_{\mathrm{basic}}$ at every step. In our experiments, we use $e_{\mathrm{act}} = \num{2e-5}$ and $e_{\mathrm{basic}} = 0.001$, leading to the consumption of $0.002 \sim 0.003$ energy by one step. It means that agents need to eat food at least once in $500$ or $1000$ steps to maintain their energy level.

When an agent makes a child, a new agent is placed in a random location sampled from a Gaussian distribution centered around its parent, of which the standard deviation is about $8\%$ of the environmental width. We sample $10$ locations each time, and reproduction succeeds if a location doesn't conflict with other objects. Thus, reproduction is more likely to succeed in a sparse place. The child inherits a proportion of the parent's energy $\eta e$, and the parent's energy level decreases to $(1-\eta)e$, where $\eta \in [0, 1]$ is the ratio of energy sharing. We used $\eta = 0.4$ in our experiments. The child also inherits its reward function from the parent with some mutation, as described later.

Since we need many agents for evolution, the multi-agent interaction can be a huge bottleneck in our simulation. To overcome this challenge, we implement our environment using JAX Python library \citep{jax2018github} so that the entire simulation loop is executed on GPU. Inspired by recent works on 3D rigid body physics simulation using JAX (e.g., \citet{brax2021github} and MuJoCo \citep{todorov2012mujoco} MJX\footnote{\url{https://mujoco.readthedocs.io/en/stable/mjx.html}}), we implement our 2D physics engine using JAX and build our environment on top of that, optimizing it for multi-agent setting. Our simulator implements projected Gauss-Seidel method with position correction \citep{catto2005iterative} that is pretty common in 2D game physics engines such as Box2D\footnote{\url{https://box2d.org}} and Chipmunk\footnote{\url{https://chipmunk-physics.net}}. We plan to release our simulation framework as open source after publication.

\begin{figure}[t]
  \centering
  \includegraphics[width=8cm]{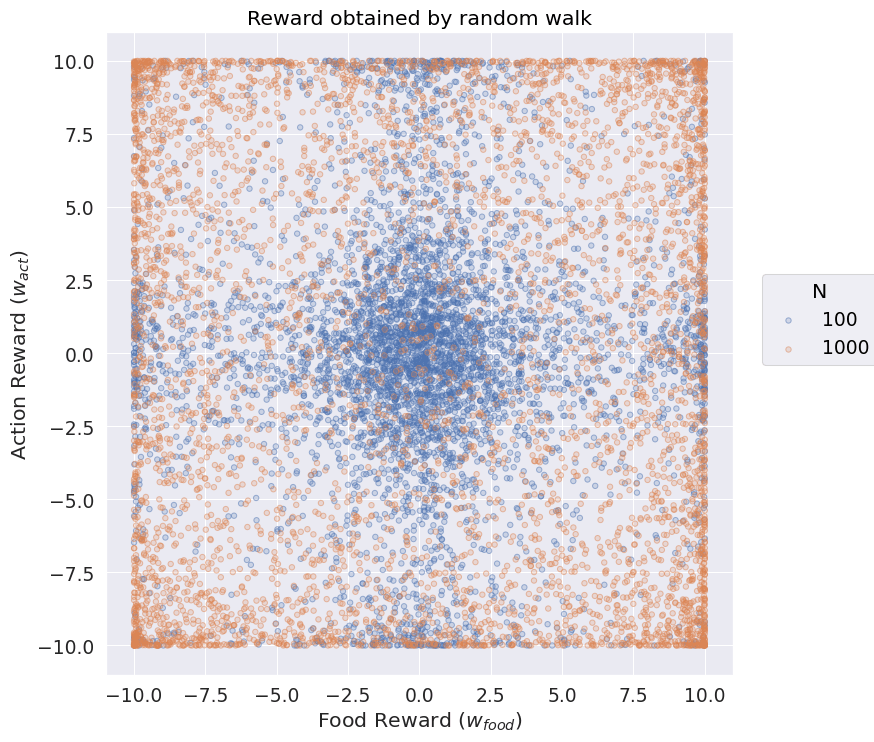}
  \caption{Reward paramters obtained by random walk with $100$ or $1000$ trials.}\label{figure:rew-random}
\end{figure}

\paragraph{Reward Function with Evolving Weights}
We assume that the reward function is determined at birth with mutation and does not change during the agent's lifetime. We use food intake and the magnitude of the agent's action (motor output) as inputs to the reward function. We expect a positive food reward to evolve to acquire energy and a negative reward for action to evolve to save energy.
We model the reward of an agent $i$ at time $t$ by:
\begin{align}
  r^{i}_{t} = w_{\mathrm{food}}^{i}n_{t}^{i} + c_\mathrm{act} w_{\mathrm{act}}^{i}|a_{t}^{i}|\label{eq:rew}.
\end{align}
In \cref{eq:rew}, $n_{t}^{i}$ is the number of foods that the agent at that step, and $w_{\mathrm{food}}^{i}$ and $w_{\mathrm{act}}^{i}$ are evolvable reward parameters of an agent $i$. Because an agent gets the action reward at every step but doesn't get the food reward so often, we use a fixed parameter $c_\mathrm{act}$ to scale the reward. We use $c_\mathrm{act}=0.01$ in our experiments. In experiments with poor or poisonous foods, we use another evolvable reward parameter $w_{\mathrm{poor}}^{i}$ or $w_{\mathrm{poison}}^{i}$.

The newborn inherits reward parameters $w_{\mathrm{food}}$ and $w_{\mathrm{act}}$ from its parent with some mutation added. To allow large jumps, we independently sample mutation noise from the Cauchy distribution with the scale parameter $0.02$ for each reward weight. After adding the Cauchy noise, each parameter is clipped to the range $[-10, 10]$ to prevent too large jumps. To clarify the characteristics of this mutation operator, we show the reward parameters $w_{\mathrm{food}}$ and $w_{\mathrm{act}}$ obtained by $100$ or $1000$ steps of a random walk with this mutation scheme in \Cref{figure:rew-random}. It shows that parameters tend to concentrate around $(0, 0)$ first and then spread into four corners in the 2D plane.

\paragraph{Reinforcement Learning}
We use Proximal Policy Optimization \citep{schulmanProximalPolicyOptimization2017} as an RL algorithm because of its fast computation time. In addition to the touch and collision sensor inputs in \Cref{figure:env-discr}, the agent takes its angle, velocity, and energy level as an observation. The action space is a continuous two-dimensional vector corresponding to the forces applied to the left and right sides of the body, clipped to $[-20, 80]$. Following the standard practice in the literature, we use a multi-layer perception (MLP) with hyperbolic tangent activation to represent the agent's policy, which is randomly initialized for each newborn. We show RL parameters used in our experiments in \Cref{tab:rl-param}.

% Add a line for reward acquisition; after observation or action?
\begin{algorithm}[t]
  \caption{Reward evolution with asexual reproduction}\label{alg:reward-evo}
  \begin{tabular}{lll}
    \textbf{Input:} & $Pop$ & Initial population of agents \\
                    & $Env$ & Simulation environment \\
                    & $h, b$ & Hazard and birth functions \\
                    & $N$ & Rollout step used in RL \\
                    & $\eta$ & Energy share ratio
  \end{tabular}
  \begin{algorithmic}[1]
    \Loop{}
    \LComment{Interact with environment}
    \ForAll{$agent \in Pop$}
      \State{$o \gets agent$'s observation in $Env$}
      \State{$a \gets $ sample from $agents$'s policy $\pi_{agent}(\cdot|o)$}
      \Once{in $N$ steps}
        \State{Update $agent$'s policy $\pi_{agent}$ via RL}
      \EndOnce{}
    \EndFor{}
    \State{Step $Env$ using collected actions}
    \State{Update $agent$'s energy level}
    \LComment{Process birth and death}
    \ForAll{$agent \in Pop$ with energy $e$ and age $t$}
      \State{$e \gets e + e_{\mathrm{food}} n_{\mathrm{food}} - e_{\mathrm{act}}|a| - e_{\mathrm{basic}}$}
      \With{Probability $b(e)$} \Comment{Birth}
        \State{$e \gets (1 - \eta) e$}
        \State{$w_{\mathrm{food}}, w_{\mathrm{act}} \gets agent$'s reward parameters}
        \State{$\delta_{\mathrm{food}}, \delta_{\mathrm{act}} \gets $ Cauchy noise}
        \State{$w_{\mathrm{food}}' \gets \max(\min(w_{\mathrm{food}} + \delta_{\mathrm{food}}, 10), -10)$}
        \State{$w_{\mathrm{act}}' \gets \max(\min(w_{\mathrm{act}} + \delta_{\mathrm{act}}, 10), -10)$}
        \State{Create a new agent with $\eta e$, $w_{\mathrm{food}}$ and $w_{\mathrm{act}}$}
      \EndWith{}
      \With{Probability $h(t, e)$} \Comment{Death}
        \State{$agent$ is removed from $Pop$ and $Env$}
      \EndWith{}
    \EndFor{}
  \EndLoop{}
\end{algorithmic}
\end{algorithm}

\begin{table}[t]
  \centering
  \begin{tabular}{ll}
    \toprule
    Parameter & Value \\
    \midrule
    Discount factor ($\gamma$) & 0.999 \\
    Rollout steps ($N$) & 1024 \\
    Minibatch size & 256 \\
    Number of optimization epochs & 10 \\
    PPO Clipping parameter & 0.2 \\
    Entropy coeff. & 0.0 \\
    GAE parameter ($\gamma$) & 0.95 \\
    Adam learning rate & \num{3e-4} \\
    Adam $\epsilon$ & \num{1e-7} \\
    Size of hidden layer in MLP & 64 \\
    \bottomrule
  \end{tabular}
  \caption{RL parameters}\label{tab:rl-param}
\end{table}

\paragraph{Simulation Procedure}
We show the pseudocode of our evolutionary simulation procedure in \Cref{alg:reward-evo}. Each agent has reward parameters $w_{\mathrm{food}}$, $w_{\mathrm{act}}$, and MLP policy. At each step, it observes sensory inputs from the environment and takes action using the MLP policy. Once in $N$ steps, the agent updates its policy via RL using the past $N$ step experiences. After environmental interaction, we evaluate all agents using the birth function $b$ and hazard function $h$ to decide whether the agent makes a child or dies. The new agent inherits a fraction of the parent's energy and mutated reward function.

\begin{table}[t]
  \centering
  \begin{tabular}{llc}
    \toprule
    $e_{\mathrm{basic}}$ & $e_{\mathrm{act}}$ & Extinction rate \\
    \midrule
    $0.002$ & $\num{2e-5}$ & $90\%$ \\
    $0.0015$ & $\num{2e-5}$ & $30\%$ \\
    $0.001$ & $\num{4e-5}$ & $100\%$ \\
    $0.001$ & $\num{3e-5}$ & $50\%$ \\
    $0.001$ & $\num{2e-5}$ & $0\%$ \\
    \bottomrule
  \end{tabular}
  \caption{Comparison of extinction rate with different metabolic parameters, measured by ten runs with different random seeds. Based on this result, $e_{\mathrm{basic}} = 0.001$ and $e_{\mathrm{act}} = \num{2e-5}$ is chosen in our experiments.}\label{table:er}
\end{table}

\begin{figure}[t]
  \centering
  \includegraphics[width=8cm]{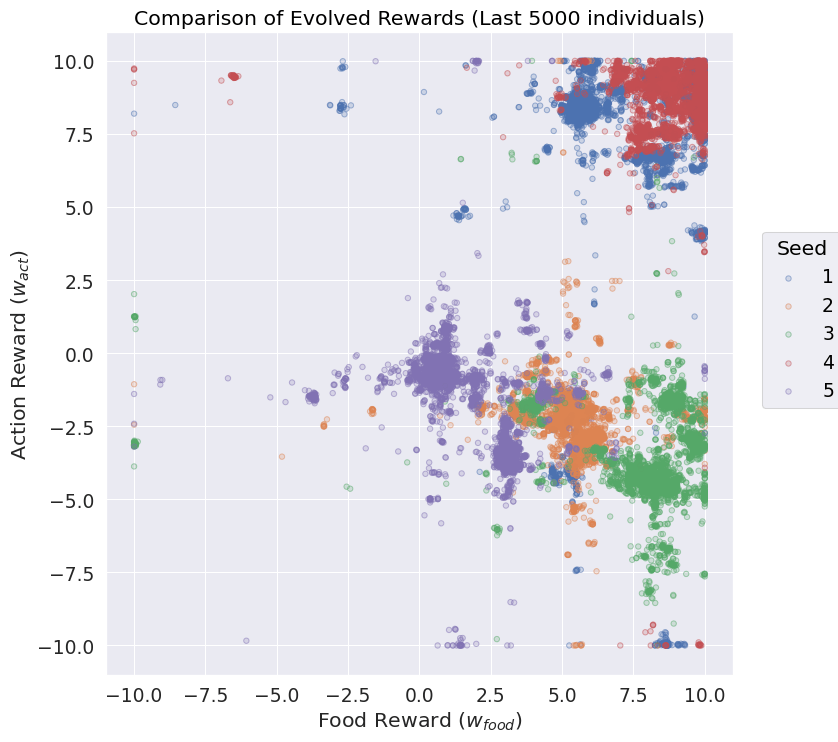}
  \caption{
    Evolved rewards of the last $5000$ agents in the baseline setting.
    The x-axis shows the weight for the food reward ($w_{\mathrm{food}}$), and the y-axis shows the weight for the action reward ($w_{\mathrm{act}}$).
    Different colors correspond to runs with different random seeds.
  }\label{figure:result-baseline}
\end{figure}

\section{Results}
In this section, we show our simulation results. We first show that biologically plausible reward functions evolve in our framework and then examine the effect of several environmental conditions, including food density and distribution. Furthermore, we conduct experiments with poor or poisonous foods and analyze their effect on evolved rewards. The source code is available at \url{https://github.com/oist/emevo}.

As a baseline environment, we use the setting where food is regenerated in a uniformly random location in the 2D environment, as the left figure in \Cref{figure:env} shows. All experiments start from $50$ agents with an energy level of $20.0$. Birth and hazard parameters (\Cref{table:hb}) are tuned to keep the population size around $130\sim 150$ through evolution. We set the population limit to $200$, which is about $30\%$ of the baseline environment.  Notably, metabolic parameters ($e_{\mathrm{basic}} = 0.001$ and $e_{\mathrm{act}} = \num{2e-5}$) are chosen so that agents don't get extinct and the evolution stably goes on. We show the extinction rate with slightly increased metabolism in \Cref{table:er}, measured by ten runs with different seeds up to about one million steps ($\num{1024e3}$). Because slightly increased metabolism leads to a $30\sim50\%$ extinction rate, we can say that these parameters are somewhat milder than the margin where extinction can happen.

Initial reward weights are sampled from a zero-mean Gaussian distribution with a standard deviation of $0.1$, so they should take values close to $0$. We conducted about ten million steps ($\num{1024e4}$) of the simulation, resulting in $300$ to $400$ generations. The complete simulation takes $10\sim12$ hours on a cluster node with an NVIDIA A100 GPU, which we think is pretty fast given that the simulation includes the interaction of $100\sim200$ agents.

\Cref{figure:result-baseline} shows the reward weights of the last $5000$ agents with five random seeds in the baseline environment, where the x-axis shows $w_{\mathrm{food}}$ and the y-axis shows $w_{\mathrm{act}}$. First, compared to the result by a random walk in \Cref{figure:rew-random}, the resulting reward weight is biased, so we conclude that evolution happened by our simulation. Food rewards are almost positive in all five runs; most exceed $4$. Thus, we can say that agents have acquired naturalistic positive food rewards only from birth and death rules.

On the contrary, action rewards have evolved into two different modes: largely positive (seed $1$ and $4$) or slightly negative (other seeds). The evolution of slightly negative rewards for motor actions aligns with our hypothesis and is reasonable in that it helps agents save their energy. However, it is surprising that positive rewards for taking large motor actions can also be stable. We guess that these positive rewards enhance explorative behavior, especially when agents are just born and have fewer samples for learning. It can give them good reinforcement signals to move even if they haven't eaten any food. However, positive rewards for action can also be disruptive for agents who already have learned to forage. This finding suggests the necessity of introducing children to adult development of RL parameters or rewards.

\begin{figure}[t]
  \centering
  \includegraphics[width=8cm]{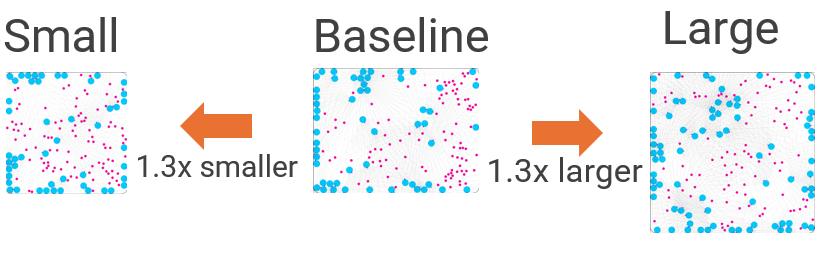}
  \caption{
    Three different sizes of environments in our experiments. Small is $360\times360$, baseline is $480\times360$, and large is $480\times480$.
  }\label{figure:envsize}
\end{figure}

\begin{figure}[t]
  \centering
  \includegraphics[width=8.4cm]{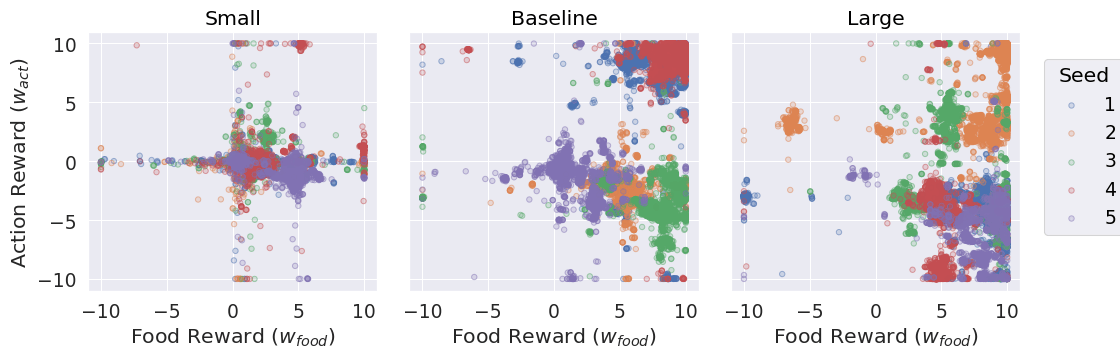}
  \caption{
    Comparison of evolved rewards of the last $5000$ agents with small, baseline, and large environment sizes from left to right.
    The x-axis shows the weight for the food reward ($w_{\mathrm{food}}$), and the y-axis shows the weight for the action reward ($w_{\mathrm{act}}$).
    Different colors correspond to runs with different random seeds.
  }\label{figure:result-envsize}
\end{figure}

\begin{table}[t]
    \centering
    \begin{tabular}{lcc}
      \toprule
      Env. Size &  Avg. Lifetime & Avg. Food Consumption \\
      \midrule
      Small &  120668 & 0.0045 \\
      Baseline & 6310 & 0.001 \\
      Large & 5718 & 0.0008 \\
      \bottomrule
    \end{tabular}
  \caption{
    Comparison of agents' average lifetime and food consumption per step in three environments with different sizes. We can see that the small environment is significantly easier, where the average lifetime of agents is much longer, and agents eat much more foods.
  }\label{table:envsize}
\end{table}

\begin{figure}[ht]
  \centering
  \includegraphics[width=6cm]{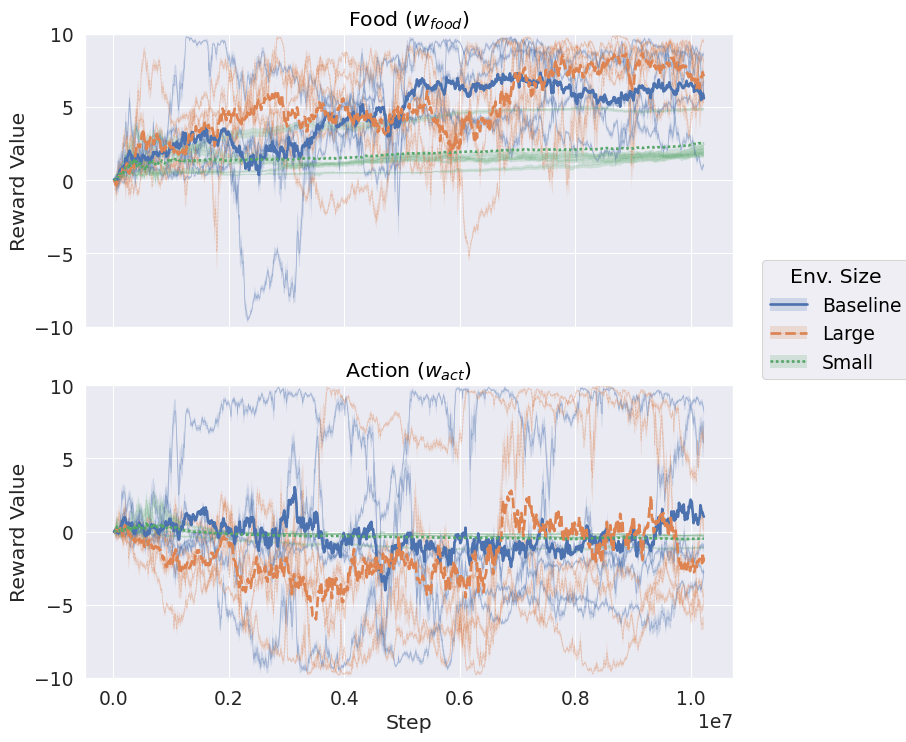}
  \caption{
    Changes in reward weights over time.
    Different colors indicate different environmental sizes.
    The upper row shows the reward weight for food ($w_{\mathrm{food}}$), and the lower row shows the reward weight for action ($w_{\mathrm{act}}$).
    Each line shows the reward weight averaged over all agents alive at that time, while thin lines show the averages within one random seed.
  }\label{figure:result-envsize-dyn}
\end{figure}

\paragraph{Effect of Food Density}
We conduct experiments with different food densities to see how environmental affordability affects the evolved rewards. With the same food regeneration parameters ($n_{\mathrm{max}} = 100$ and $g = 0.02$), we use three environmental sizes: small, baseline, and large, as shown in \Cref{figure:envsize}. We show the evolved rewards with these three different environment sizes in \Cref{figure:result-envsize}. Foods are regenerated more densely in smaller environments. Rewards evolved in a large environment show characteristics similar to the baseline. I.e., food rewards are consistently positive, and action rewards are positive or negative. However, action rewards tend to be slightly smaller than the baseline, which may reflect the importance of saving energy in this environment. Actually, as \Cref{table:envsize} shows, the agent's average lifetime and food consumption are the least in the large environment.

Rewards in the small environment show different properties from the ones that evolved in the other two environments. Food reward is slightly positive but rarely becomes large. We don't see any significant tendency in the weight for action rewards. We guess this is because it's too easy for agents to get food in this environment, and the selection pressure is much weaker. As we show in \Cref{table:envsize}, the agent's average lifetime and food consumption are much higher in this environment. This tendency is also reflected in the change of reward weights over time shown in \Cref{figure:result-envsize-dyn}. We can see that the evolution of reward parameters is by far the most stable in the small environment. On the contrary, reward parameters change in the baseline and large environments. We can especially see large jumps in action rewards. This result suggests that limiting food resources is necessary for the rapid evolution of food rewards, and too much food supply can stop evolution.

\begin{figure}[t]
  \centering
  \includegraphics[width=8cm]{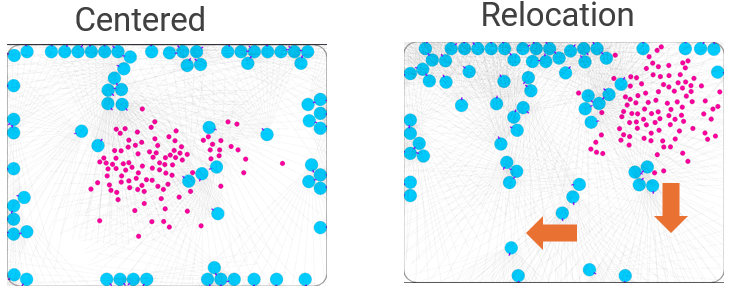}
  \caption{
    Environments with centered food distribution (left) and relocating food locations (right).
  }\label{figure:env-fdist}
\end{figure}

\begin{figure}[t]
  \centering
  \includegraphics[width=8.4cm]{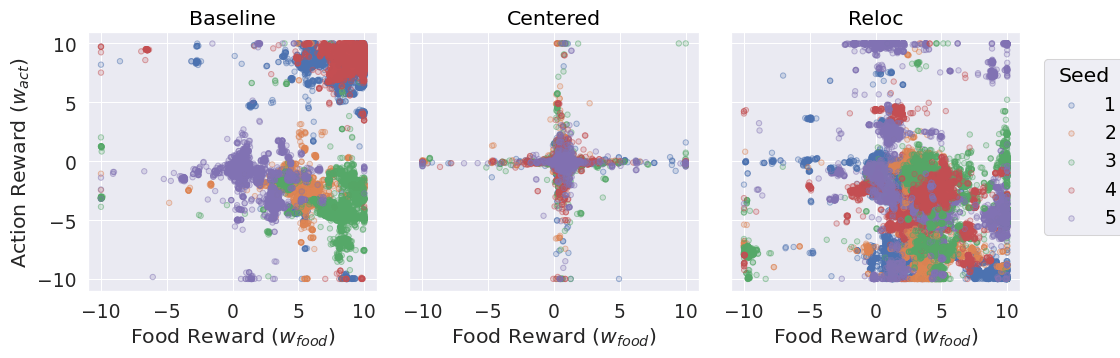}
  \caption{
    Comparison of evolved rewards of the last $5000$ agents with the baseline environment, the environment with centered food distribution, and the environment with food relocation from left to right.
    The x-axis shows the weight for the food reward ($w_{\mathrm{food}}$), and the y-axis shows the weight for the action reward ($w_{\mathrm{act}}$).
    Different colors correspond to runs with different random seeds.
  }\label{figure:result-fdist}
\end{figure}

\begin{table}[t]
    \centering
    \begin{tabular}{lcc}
      \toprule
      Food Dist. &  Avg. Lifetime & Avg. Food Consumption \\
      \midrule
      Baseline & 6310 & 0.001 \\
      Centered &  74318 & 0.0039 \\
      Relocation & 6794 & 0.00096 \\
      \bottomrule
    \end{tabular}
  \caption{
    Comparison of agents' average lifetime and food consumption per step in environments with different food distributions.
  }\label{table:fdist}
\end{table}

\paragraph{Effect of Food Distribution}
In addition to the food density, how foods are distributed over the environment can also affect the rewards. It may be possible that uniformly sampling food locations makes it easy for agents who move randomly to survive in the environment, so the bias in food location can make RL more important. To confirm this hypothesis, we conduct experiments with different food distributions as shown in \Cref{figure:env-fdist}. In an environment with a centered food distribution, foods are sampled from the Gaussian distribution around the center of the environment. We also use the environment with food relocation, where foods are sampled from the Gaussian as in the centered' environment, but the center of the Gaussian moves clockwise between each corner of the environment. This food relocation periodically happens after $1000$ foods are eaten. The same food regeneration parameters ($n_{\mathrm{max}} = 100$ and $g = 0.02$) are used.

We show the evolved rewards of the last $5000$ agents in these environments in \Cref{figure:result-fdist}. We can see that almost nothing has evolved in the centered environment, which suggests that it is too easy for agents to get food in this environment. However, as \Cref{table:fdist} shows, the agent's average lifetime and food consumption are less than the small environment, so we can say that this environment with centered food distribution is easy but slightly more difficult than the small environment. Thus, we conclude that the food and action rewards are less likely to evolve in this environment with the centered distribution of foods than in an environment with uniformly distributed foods. We speculate that this is because the place where agents are born can be a significant bias for survival, making rewards less important.

On the contrary, we can see that reward parameters have evolved similarly to the baseline in the environment with food relocation, and the reward for motor action is mostly negative in all five runs with different seeds. To check whether the difference in difficulty causes this shift of action rewards, we show the average lifetime and food consumption in the baseline and the food relocation environment in \Cref{table:fdist}, where we find that their difficulties are very close in. Thus, we argue that this result complements our hypothesis that the bias of the food locations makes RL more important, leading to more energy-efficient rewards.

\begin{table}[t]
  \begin{subtable}[h]{0.45\columnwidth}
    \centering
    \begin{tabular}{cc}
      \toprule
      Parameter & Value \\
      \midrule
      $n_{\textrm{max}}^{\textrm{normal}}$ & 90 \\
      $n_{\textrm{max}}^{\textrm{poor}}$ & 30 \\
      \bottomrule
    \end{tabular}
  \end{subtable}
  \begin{subtable}[h]{0.45\columnwidth}
    \centering
    \begin{tabular}{cc}
      \toprule
      Parameter & Value \\
      \midrule
      $n_{\textrm{max}}^{\textrm{normal}}$ & 120 \\
      $n_{\textrm{max}}^{\textrm{poison}}$ & 40 \\
      \bottomrule
    \end{tabular}
  \end{subtable}
  \caption{
    Food regeneration parameters with poor foods (left) or poisonous foods (right).
    $n_{\textrm{max}}^{\textrm{poor}}$ and $n_{\textrm{max}}^{\textrm{poison}}$ are the maximum number of poor or poisonous foods, while $g^{\textrm{poor}}$ and $g^{\textrm{poison}}$ are growth rate of poor or poisonous foods.
  }\label{table:pp}
\end{table}

\begin{figure}[t]
  \begin{subfigure}[t]{4cm}
    \centering
    \includegraphics[width=4cm]{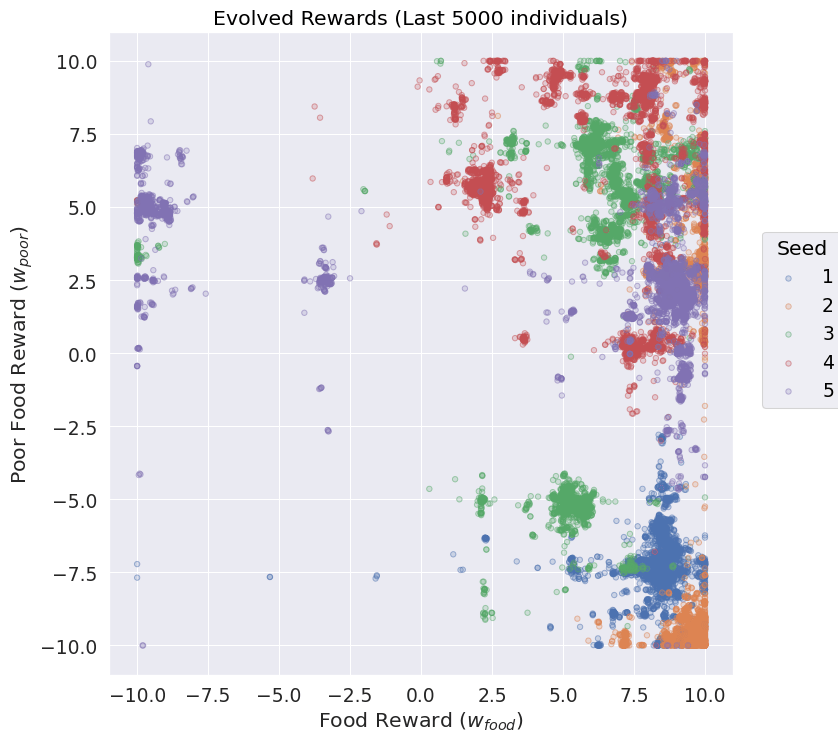}
  \end{subfigure}
  \begin{subfigure}[t]{3.5cm}
    \centering
    \includegraphics[width=4cm]{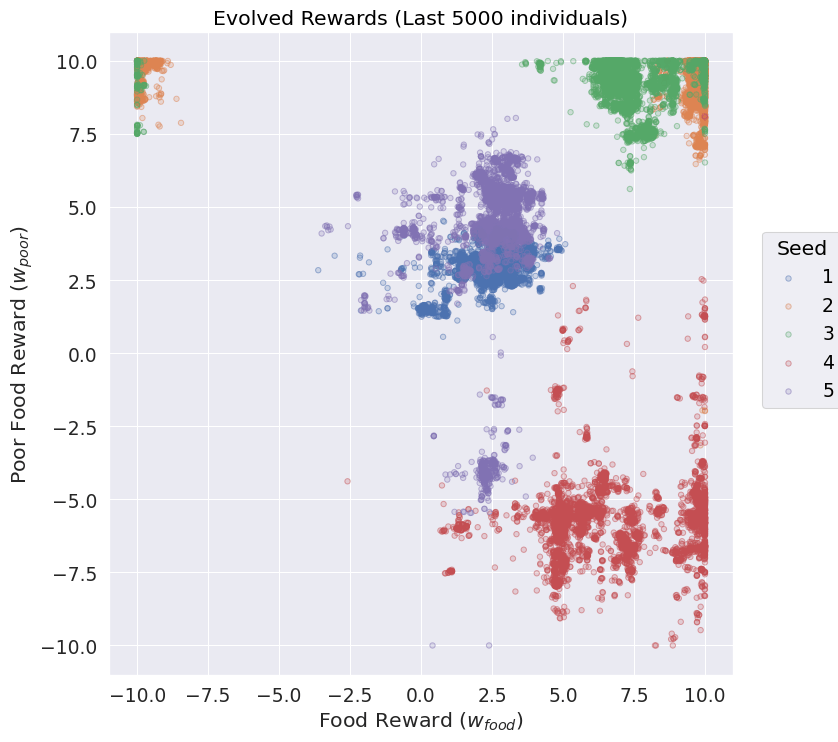}
  \end{subfigure}
  \caption{
    \textbf{Left:} Evolved rewards of the last $5000$ agents with poor foods, where the x-axis shows $w_{\mathrm{food}}$, and the y-axis shows $w_{\mathrm{poor}}$.
    \textbf{Right:} Evolved rewards with poisonous foods, where the x-axis shows $w_{\mathrm{food}}$, and the y-axis shows $w_{\mathrm{poison}}$.
    Different colors correspond to runs with different random seeds.
  }\label{figure:result-pp}
\end{figure}

\paragraph{Effect of Poor and Poisonous Foods}
In the experiments, only one kind of food with $e_{\mathrm{food}} = 1.0$ in the environment. However, one role of our reward system is to distinguish between different external stimuli. For example, our food reward system helps us eat nutritious food and avoid poisonous foods. Therefore, we conduct simulations with poor and poisonous foods, of which the energy gains are $e_{\mathrm{poor}} = 0.2$ and $e_{\mathrm{poison}} = -0.6$. In those environments, both normal and poor or poisonous foods are sampled uniformly over the entire environment. The number of maximum foods is smaller in the environment with poor foods, and there are more normal foods in the environment with poisonous foods, as shown in \Cref{table:pp}, so the overall energy supply should be roughly the same.

\begin{figure}[t]
  \centering
  \includegraphics[width=6cm]{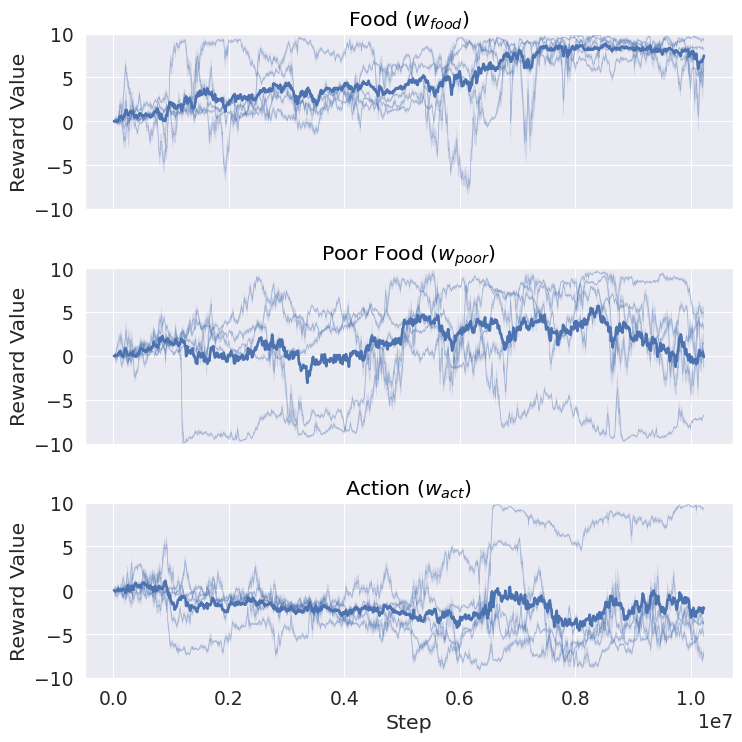}
  \caption{
    Evolved rewards with poor foods.
    From top to bottom, each row shows the reward weight for food ($w_{\mathrm{food}}$), the reward weight for poor food ($w_{\mathrm{poor}}$), and the reward weight for action ($w_{\mathrm{act}}$).
    Each line shows the reward weight averaged over all agents alive at that time, while thin lines show the averages within one random seed.
  }\label{figure:result-dyn-poor}
\end{figure}

The left panel in \Cref{figure:result-pp} shows the evolved reward weights with poor foods, where the x-axis shows $w_{\mathrm{food}}$, and the y-axis shows $w_{\mathrm{poor}}$. Interestingly, while rewards for normal foods are always largely positive, rewards for poor foods are spread from mainly negative to positive. This is a surprising result because agents still gain energy by eating poor foods, and they don't need to avoid them. We suspect that such negative rewards for foods can be useful in prioritizing hunting for normal foods. This tendency is also seen in the rewards change over time, as shown in \Cref{figure:result-dyn-poor}. We can see that rewards for poor foods are unstable and diverging in various directions, while rewards for normal foods converge.

\begin{figure}[t]
  \centering
  \includegraphics[width=6cm]{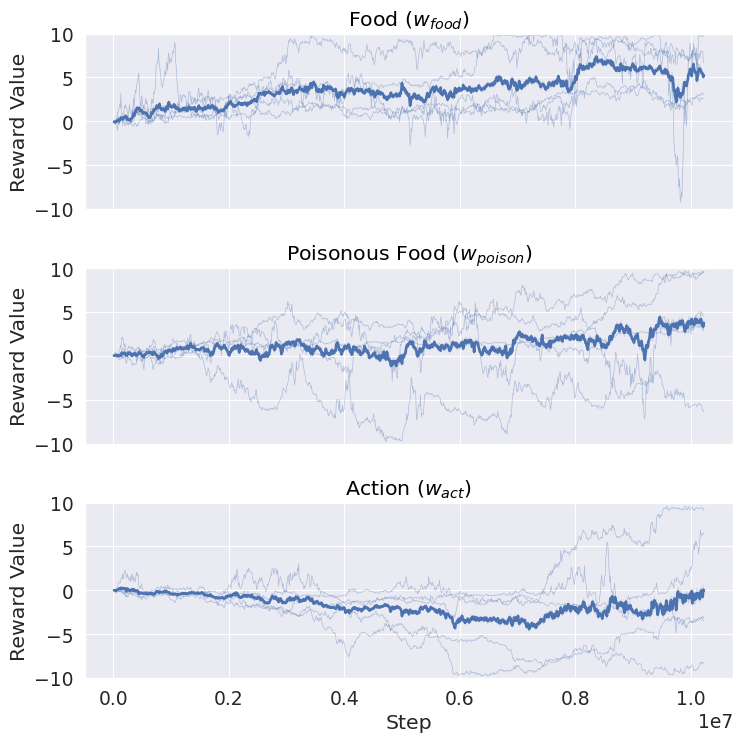}
  \caption{
    Evolved rewards with poisonous foods.
    From top to bottom, each row shows the reward weight for food ($w_{\mathrm{food}}$), the reward weight for poisonous food ($w_{\mathrm{poison}}$), and the reward weight for action ($w_{\mathrm{act}}$).
  }\label{figure:result-dyn-poison}
\end{figure}

More surprisingly, the same tendency is seen in the environment with poisonous foods. The right panel in \Cref{figure:result-pp} shows the evolved reward weights with poisonous foods, where the x-axis shows $w_{\mathrm{food}}$, and the y-axis shows $w_{\mathrm{poison}}$. We can see that rewards for poisonous foods are positive in most random seeds, though agents lose their energy by eating poisonous foods. It's intriguing that they still can survive with positive poison rewards, although they have larger rewards for normal foods. The dynamics of reward weights shown in \Cref{figure:result-dyn-poison} is also quite similar to the one in the environment with poor foods: rewards for poisonous foods and actions diverge. In contrast, the reward for normal foods converges to largely positive value. Again, this result suggests that less-important reward signals can have a wide variety, like our preference, while the primal reward signal is very stable.

\section{Conclusion}
In this paper, we propose a distributed evolution framework with a simple birth and death model to study the evolution of reward function. Our results show that biologically reasonable reward functions can evolve from randomly initialized ones. We also find that environmental conditions, including food density and relocation, can affect rewards. Notably, we find that too much food supply or too strong of a bias can stop evolution, and less food supply or bias in food location can lead to fewer action rewards. Moreover, we examine the evolution of rewards for poor and poisonous rewards, finding that rewards for such less important foods can be unstable, while rewards for normal foods are stably positive. These results suggest that the evolution of reward function is led by one nominal reward, in our case, the reward for normal food. The other reward components can be unstable and diverge in many directions, even if they benefit survival.  An interesting future direction is to see the effect of multi-species relationships like predator and prey, which can be a stronger motivation to evolve better rewards for competition. Relationships with homeostasis-based learning \citep{yoshidaEmergenceIntegratedBehaviors2024} and the development of internal parameters may also be interesting.

%\subsubsection*{Acknowledgments}\label{sec:ack}
% Use unnumbered third level headings for the acknowledgments. All acknowledgments, including those to funding agencies, go at the end of the paper. Only add this information once your submission is accepted and deanonymized.

%%%%%%%%%%%%%%%%%%%%%%%%%%%%%%%%%%%%%%%%%%%%%%%%%%%%%%%%%%%%%%%%
%% Bibliography
%%%%%%%%%%%%%%%%%%%%%%%%%%%%%%%%%%%%%%%%%%%%%%%%%%%%%%%%%%%%%%%%
\clearpage
\bibliographystyle{apalike}
\bibliography{references}

%%%%%%%%%%%%%%%%%%%%%%%%%%%%%%%%%%%%%%%%%%%%%%%%%%%%%%%%%%%%%%%%
%% Appendices
%%%%%%%%%%%%%%%%%%%%%%%%%%%%%%%%%%%%%%%%%%%%%%%%%%%%%%%%%%%%%%%%

% Retrieve appendix for arxiv?
% \clearpage
% \appendix
% \input{src/appendix.tex}

\end{document}